\documentclass[runningheads,orivec]{llncs}
\usepackage[T1]{fontenc}

\usepackage{graphicx}
\usepackage{booktabs}
\usepackage{xcolor}

\usepackage{hyperref} 
\usepackage{comment}
\usepackage{times}
\usepackage{bibunits}

\usepackage{makecell}
\usepackage{multirow}
\usepackage{amssymb}
\usepackage{xcolor}

\usepackage{amsmath}
\usepackage[ruled,vlined]{algorithm2e}
\SetKwInput{KwInput}{Input}                
\SetKwInput{KwOutput}{Output}              

\definecolor{codegreen}{rgb}{0,0.6,0}
\definecolor{codegray}{rgb}{0.5,0.5,0.5}
\definecolor{codepurple}{rgb}{0.58,0,0.82}
\definecolor{backcolour}{rgb}{0.95,0.95,0.92}

\usepackage{xcolor}

\newcommand{\modelname}{\texttt{CogGNN}\xspace}
\newcommand{\submodelname}{\texttt{Vis-CogGNN}\xspace}

\usepackage{tikz,xcolor,hyperref}

\definecolor{lime}{HTML}{A6CE39}
\DeclareRobustCommand{\orcidicon}{
	\begin{tikzpicture}
	\draw[lime, fill=lime] (0,0) 
	circle [radius=0.16] 
	node[white] {{\fontfamily{qag}\selectfont \tiny ID}};
	\draw[white, fill=white] (-0.0625,0.095) 
	circle [radius=0.007];
	\end{tikzpicture}
	\hspace{-2mm}
}
\foreach \x in {A, ..., Z}{\expandafter\xdef\csname orcid\x\endcsname{\noexpand\href{https://orcid.org/\csname orcidauthor\x\endcsname}
			{\noexpand\orcidicon}}
}
			
\usepackage{tikz}
\usepackage{xparse}   

\DeclareRobustCommand{\authorpic}[2][5mm]{%
  \tikz[baseline={([yshift=-.25ex]current bounding box.center)}]{%
    \clip (0,0) circle (#1);
    \pgfmathsetlengthmacro{\picside}{sqrt(2)*#1}%
    \node at (0,0) {\includegraphics[width=\picside,height=\picside,keepaspectratio]{#2}};
    \draw[line width=0.4pt, color=white] (0,0) circle (#1);
  }%
}

\NewDocumentCommand{\AuthorWithPic}{O{5.5mm} O{0.20em} m m}{%
  \texorpdfstring{\authorpic[#1]{#4}\kern #2}{}%
  #3%
}

\usepackage{xcolor}
\definecolor{linkpinkix}{HTML}{EA335A} 

\definecolor{linkpink}{HTML}{EA335A}

\newcommand{\shadedlink}[2]{%
  \tikz[baseline=(n.base)]\node[
    fill=linkpink,
    fill opacity=0.5,
    text opacity=1,
    rounded corners=.3ex,
    inner xsep=.35em,
    inner ysep=.15em
  ] (n) {\href{#1}{\textcolor{blue!70!black}{#2}}};%
}

\begin{document}

\title{\modelname: Cognitive Graph Neural Networks in Generative Connectomics}
\author{
\AuthorWithPic[6mm][0.18em]{Mayssa Soussia}{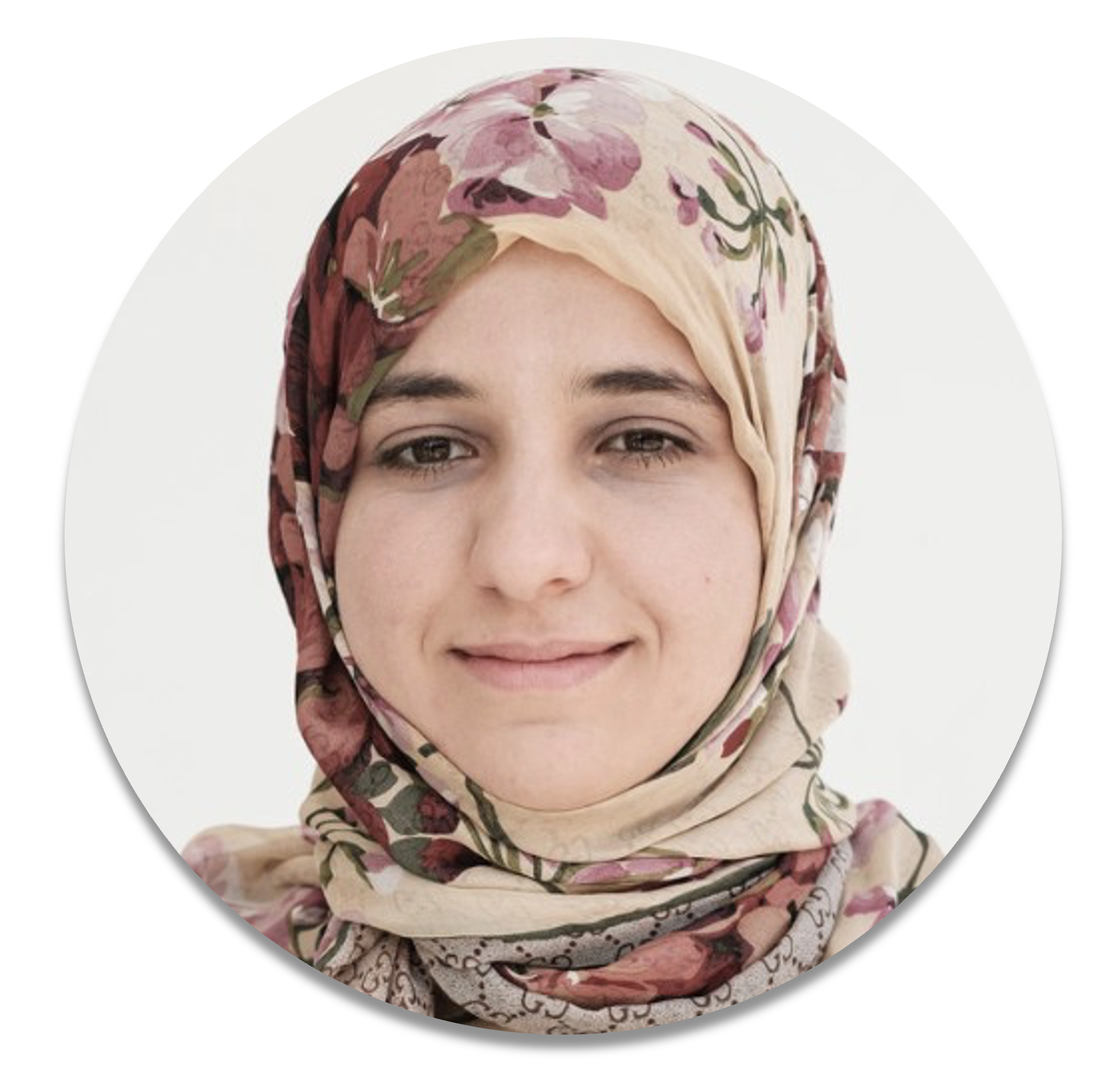}\inst{1,2} \and
\AuthorWithPic[6mm][0.18em]{Yijun Lin}{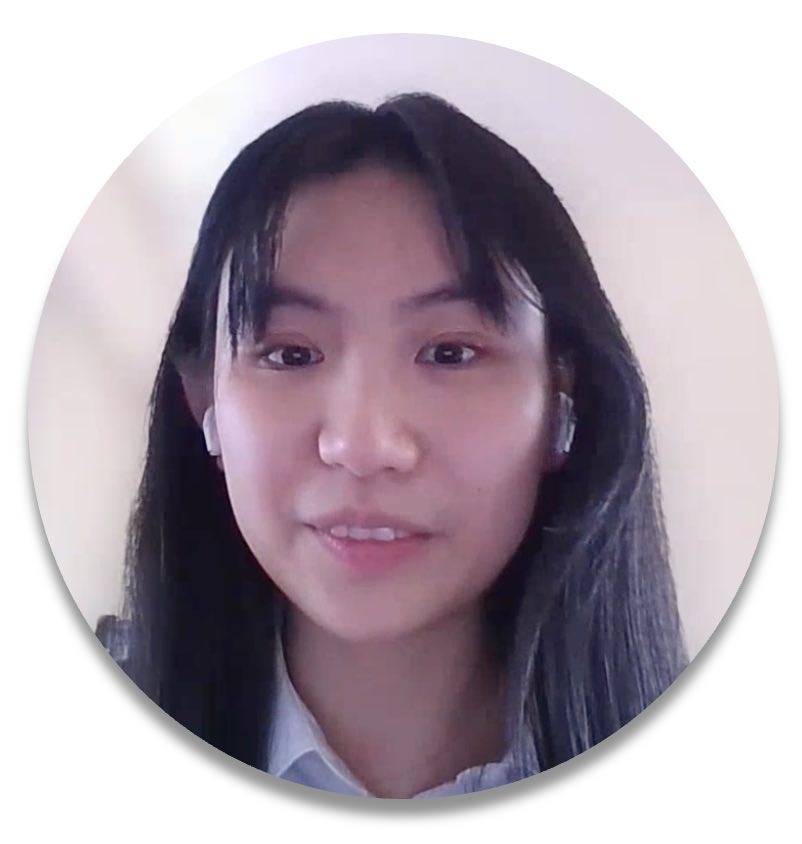} \inst{2} \and
\AuthorWithPic[6mm][0.18em]{Mohamed Ali Mahjoub}{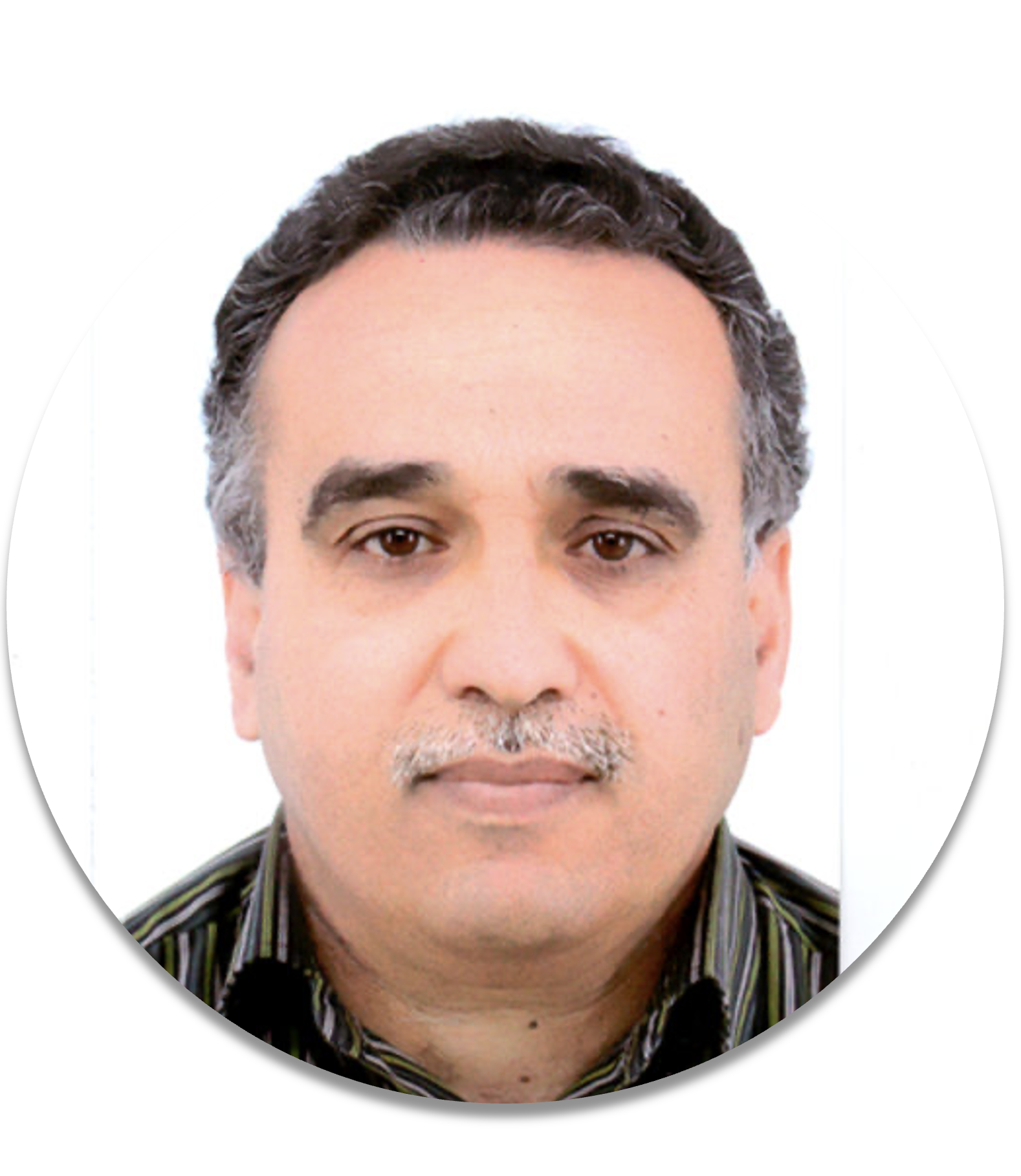}\inst{1} \and
\AuthorWithPic[6mm][0.18em]{Islem Rekik}{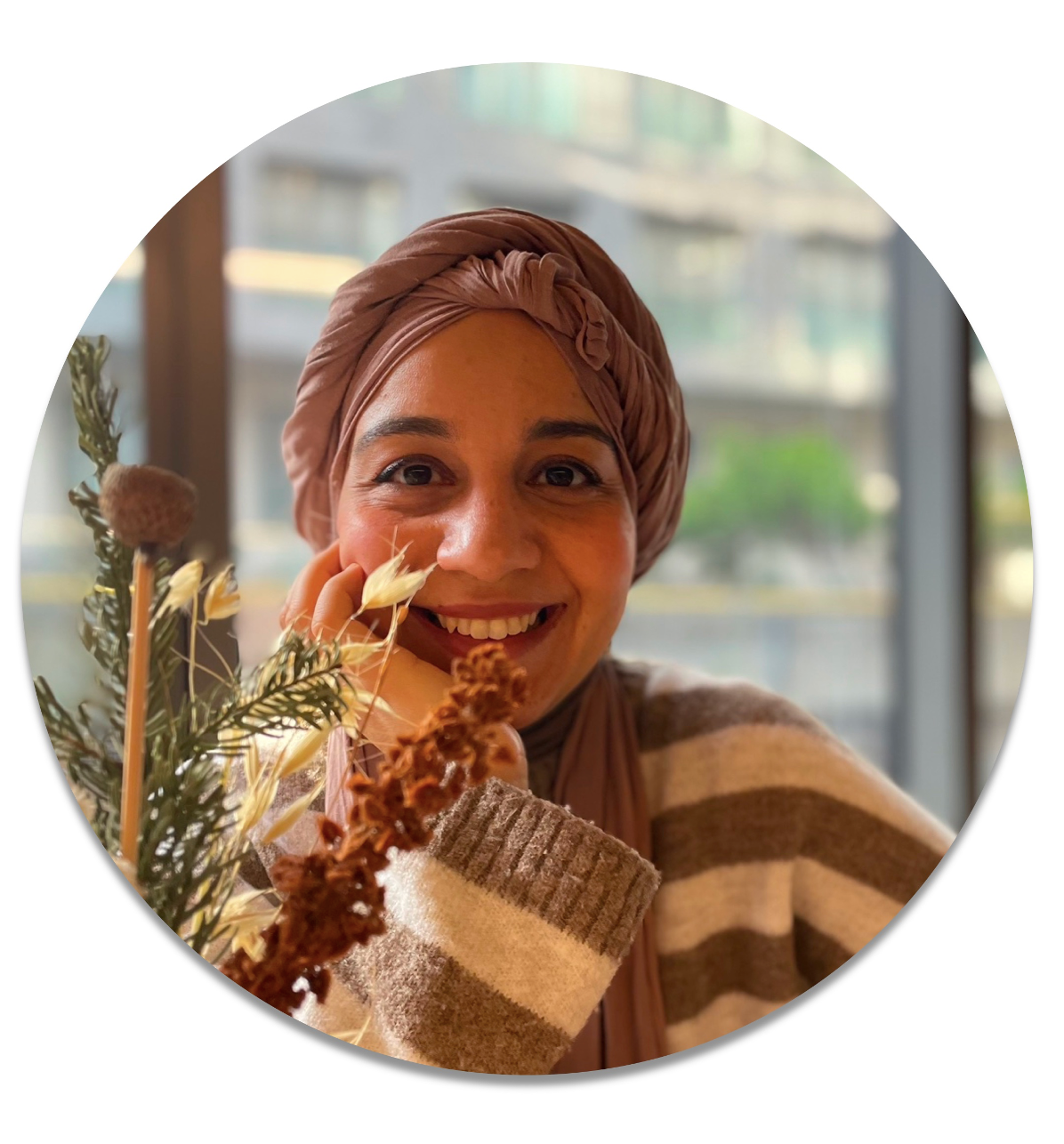}\inst{2}\orcidA{} \thanks{ {corresponding author: i.rekik@imperial.ac.uk, 
\url{http://basira-lab.com}. }}}

\authorrunning{Soussia et al.}
%
\institute{National Engineering School of Sousse, University of Sousse, LATIS- Laboratory of Advanced Technology and Intelligent Systems, 4023, Sousse, Tunisia\and
BASIRA Lab, Imperial-X and Department of Computing, Imperial College London, UK  \\
 }
\maketitle              
\begin{abstract}
Generative learning has advanced network neuroscience, enabling tasks like graph super-resolution, temporal graph prediction, and multimodal brain graph fusion. However, current methods, mainly based on graph neural networks (GNNs), focus solely on structural and topological properties, neglecting cognitive traits. To address this, we introduce the first \emph{cognified} generative model, \textsc{\modelname}, which endows GNNs with cognitive capabilities (e.g., visual memory) to generate brain networks that preserve cognitive features. While broadly applicable, we present \textsc{\submodelname}, a specific variant designed to integrate visual input, a key factor in brain functions like pattern recognition and memory recall. As a proof of concept, we use our model to learn connectional brain templates (CBTs), population-level fingerprints from multi-view brain networks. Unlike prior work that overlooks cognitive properties, \textsc{\submodelname} generates CBTs that are both cognitively and structurally meaningful. Our contributions are: (i) a novel cognition-aware generative model with a visual-memory-based loss; (ii) a CBT-learning framework with a co-optimization strategy to yield well-centered, discriminative, cognitively enhanced templates. Extensive experiments show that \textsc{\submodelname} outperforms state-of-the-art methods, establishing a strong foundation for cognitively grounded brain network modeling. \footnote{This paper has been selected for a \textbf{Short Oral Presentation} at the DGM4 MICCAI 2025 workshop. \shadedlink{https://youtube/wB5i_k_U2YQ}{[\modelname YouTube Video]}.}

\keywords{Cognitive learning  \and GNNs  \and CBTs \and Reservoir computing}
\end{abstract}
%
%


\section{Introduction}
The field of connectomics has recently embraced generative modeling \cite{bessadok2022graph}, particularly GNN-based approaches, which have enabled advanced applications in brain connectivity analysis, such as \emph{graph super-resolution}, where high-resolution brain graphs are generated from lower-resolution connectomes \cite{mhiri2021non}. Another emerging direction is \emph{brain evolution prediction}, which models the temporal progression of brain connectivity from a single timepoint. EvoGraphNet~\cite{nebli2020deep} addresses this by chaining graph-based generative adversarial networks (gGANs) to predict future brain graphs. \emph{Cross-modality brain graph synthesis} is also gaining attention. Here, generative models learn to produce a connectome in one modality (e.g., functional) from another (e.g., structural) . For instance, Zhang et al.~\cite{zhang2020deep} proposed a GCN that generates functional networks from structural ones. Finally, another important application is the learning of \emph{connectional brain templates (CBTs)}—population-level fingerprints of brain networks that serve as a representative connectome. Deep graph normalizer (DGN)~\cite{Gurbuz:2020} is the first GNN model designed to normalize and integrate multi-subject brain graphs into a single CBT using an end-to-end learning framework.

Despite notable advances, existing generative GNN models rely on traditional data-driven processes that capture structure and topology \textbf{but overlook cognitive capacities of brain connectomes, such as visual memory}. In this work, we introduce the first \emph{cognified} GNN model, \textsc{\modelname}, which endows generative models with cognitive abilities. We hypothesize that incorporating such capacities enables more intelligent network generation, preserving cognitive traits in brain data. While \textsc{\modelname} is a general-purpose framework, we present a focused proof of concept that integrates visual cognition into brain graph generation. Visual input plays a central role in cognition, supporting memory, pattern recognition, and perceptual integration, key processes for biologically grounded brain modeling.

To demonstrate our approach, we introduce \textsc{\submodelname}, a cognitively enhanced variant of \textsc{\modelname} that integrates visual memory into GNN-based generative learning. We focus on generating connectional brain templates (CBTs)—population, level connectomes that capture group-wise neural organization, support biomarker discovery, and enable cross-individual comparisons \cite{chaari2023comparative}. Motivated by findings that reservoir computing (RC) architectures mimic core cognitive functions of the prefrontal cortex, such as working memory, temporal integration, and flexible attention \cite{enel2016reservoir,nakamura2024reservoir}, we propose combining RC with GNNs to endow generated CBTs with memory-driven cognitive capacities. In this hybrid design, RC serves as a cognitive scaffold, mapping visual input sequences into a high-dimensional dynamic state space via a fixed recurrent network while only the output layer is trained. This architecture is both efficient and cognitively grounded, enhancing the generated brain graphs with visual memory properties.


Our contributions are fourfold: (i) \textit{Methodologically}, we introduce \textsc{\modelname}, the first cognitively enhanced GNN-based generative model that bridges structural brain modeling with high-level cognitive functions. We also propose \textsc{\submodelname}, a variant that incorporates visual capacity through a novel cognitive loss, enriching graph representations with memory-related abilities. (ii) \textsl{Application-wise}, we present a hybrid GNN-RC framework for generating connectional brain templates (CBTs) that are structurally sound, topologically meaningful, and cognitively enhanced. (iii) \textsl{Clinically}, cognitive CBTs show promise as biomarkers for assessing cognitive decline in brain disorders (e.g., Alzheimer’s disease), with potential for early diagnosis and longitudinal monitoring. (iv) \textsl{Generically}, while our focus is on visual memory, \textsc{\modelname} is adaptable to other cognitive domains (e.g., auditory or language processing) and applicable beyond CBT generation.

\begin{figure}[!ht]
    \centering
    \includegraphics[width=1\linewidth]{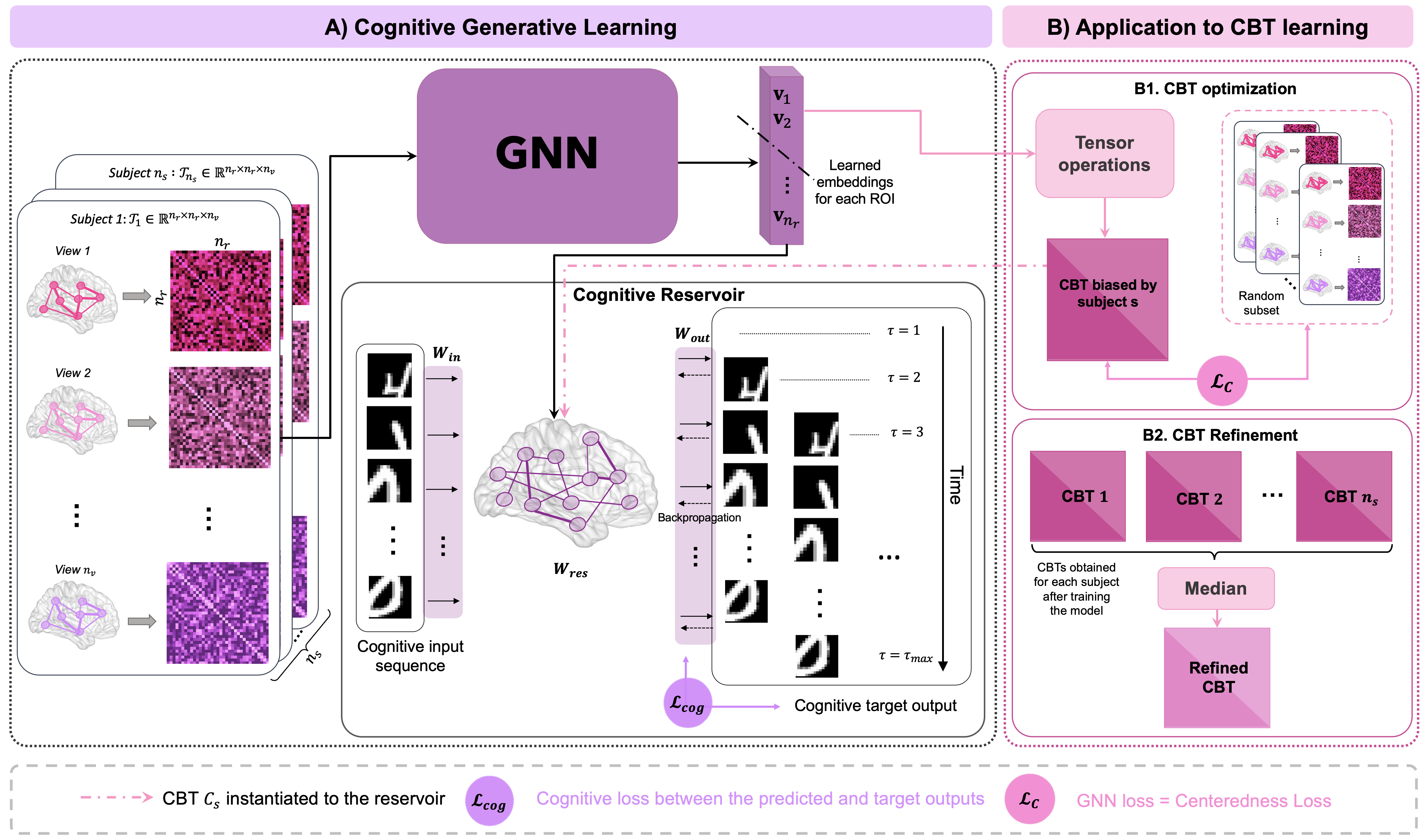}
    \caption{Illustration of \textsc{\modelname} integrating cognitive capacities into generative brain network modeling. \textbf{A) Cognitive Generative Learning}. and \textbf{B) CBT Learning} comprising (1) \textsl{CBT Optimization} and (2) \textsl{CBT Refinement}.}
    \label{fig:main}
\end{figure}
    

\section{Methods}

In this section, we present the \modelname framework, illustrated in Fig. \ref{fig:main}, consisting of two components: a generic pipeline for cognitive generative graph modeling (Fig. \ref{fig:main}.A) and a specialized application for generating CBTs (Fig.~\ref{fig:main}.B). 

\noindent\textbf{Generic Problem Statement.} Given a set of $n_s$ subjects, each represented by a tensor \( \mathcal{T}_s \in \mathbb{R}^{n_r \times n_r \times n_v} \) across \( n_v \) views and \( n_r \) regions, and a cognitive trait vector \( \mathbf{c}_s \in \mathbb{R}^{n_c} \) (e.g., visual memory), the goal is to learn a mapping $f: \{\mathcal{T}_s, \mathbf{c}_i\}_{s=1}^{n_s} \mapsto \mathbf{G}_{\text{out}} \in \mathbb{R}^{n_r \times n_r}, $
where \( \mathbf{G}_{\text{out}} \) is the generated brain graph that preserves cognitive traits.


\subsubsection{Cognitive generative learning.}The first block of \modelname (Fig \ref{fig:main}.A) consists of the generic framework designed to produce a brain graph representation that preserves a specific cognitive capacity (e.g., visual memory) and comprises two steps: 

\textsl{Multi-view graph embedding generation.} Given a set of \(n_s\) training subjects, each subject \(s\) is represented by multi-view brain networks encoded into a tensor \(\mathcal{T}_s \in \mathbb{R}^{n_r \times n_r \times n_v}\), where each view represents a specific cortical attribute (e.g., cortical thickness) (Fig. \ref{fig:main}.A). These multi-view representations are processed by a generative GNN, which captures complex interrelationships between nodes and views and generates node embeddings \(V = [\mathbf{v}_1, \mathbf{v}_2, \dots, \mathbf{v}_{n_r}]^T\) for each ROI (Fig. \ref{fig:main}.A). The choice of the GNN architecture can vary depending on the specific task requirements. The resulting node embeddings are subsequently transformed into a brain graph representation (i.e., connectome), which serves as the foundation for the reservoir computing framework. 

\textsl{Cognitive reservoir}. At the core of our framework is the \textsl{Cognitive Reservoir (CR)}, which integrates cognitive capacity into the generative learning process. We implement the CR using an Echo State Network (ESN) \cite{jaeger2001echo}, a form of reservoir computing composed of an input layer, a recurrent reservoir, and an output layer.
The CR's input layer flexibly encodes various cognitive inputs (e.g., visual, linguistic), acting as a modality-agnostic interface that projects each input \( \mathbf{c}_i \in \mathbb{R}^{n_c} \) into a high-dimensional space via an input weight matrix \( \mathbf{W}_{\text{in}} \in \mathbb{R}^{n_r \times n_c} \), where \( n_c \) is the dimensionality of the input and \( n_r \) the number of reservoir nodes (i.e., brain regions or ROIs).
The reservoir layer, or internal reservoir, is defined by a recurrent weight matrix \( \mathbf{W}_{\text{res}} \in \mathbb{R}^{n_r \times n_r} \). Inspired by the work \cite{DamicelliHilgetagGoulas2022}, we instantiate it using a real brain connectome derived from GNN-based node embeddings (Fig. \ref{fig:main}.A). This biologically grounded design enables the reservoir to function as a fixed recurrent neural network, projecting cognitive inputs into a high-dimensional dynamic space that naturally supports temporal processing and memory retention, key for modeling functions such as perception, attention, and recall.
The reservoir processes each cognitive input \( \mathbf{c}_i \) by mapping it into its dynamic state space. The reservoir state \( \mathbf{x}_i \) is updated according to:

\begin{equation}
\mathbf{x}_i = \tanh \left( \alpha \mathbf{W}_{\text{in}} [1; \mathbf{c}_i] + (1 - \alpha) \mathbf{W}_{\text{res}} \mathbf{x}_{i-1} \right),
\label{eq:state_update}
\end{equation}
where \( [1; \mathbf{c}_i] \) denotes vertical concatenation with a bias term, and \( \alpha \in [0, 1] \) controls the trade-off between the current input and the previous state. A higher \( \alpha \) emphasizes the immediate input, while a lower value gives more weight to the past, effectively extending the reservoir's memory. This tunable mechanism allows the reservoir to capture both short- and long-term dependencies in the cognitive signal stream.

The output at each time step is computed through a linear transformation of the reservoir state $\mathbf{\hat{y}}_i = \mathbf{W}_{\text{out}} \mathbf{x}_i,$
where \( \mathbf{W}_{\text{out}} \in \mathbb{R}^{n_{\text{out}} \times n_r} \) is the only trainable parameter in the system. It is optimized using a cognitive loss function \( \mathcal{L}_{\text{cog}} \), which minimizes the discrepancy between predicted outputs \( \mathbf{\hat{y}}_i \) and targets \( \mathbf{y}_i \), thereby encouraging the model to retain task-relevant cognitive features. This loss is combined with the graph-specific loss \( \mathcal{L}_{\text{gnn}} \) to define the total objective.

\subsection{\submodelname: Case-Study}
As a proof of concept, we introduce \submodelname, a specialized instance of our generic framework designed to simulate visual memory. In this setup, the cognitive input corresponds to visual stimuli, denoted as \( \mathbf{u}_i = \mathbf{I}_i \), and the model output is the predicted image \( \mathbf{\hat{y}}_i = \hat{\mathbf{I}}_i \). The primary cognitive property of interest is the model’s ability to retain and recall visual information, formally evaluated through \emph{visual memory capacity} (Vis-MC). We define a visual recall task where the reservoir is trained to reconstruct temporally delayed versions of a sequence of input images. Given a sequence \( \{ \mathbf{I}_1, \mathbf{I}_2, \dots, \mathbf{I}_{n_m} \} \), with each image \( \mathbf{I}_i \in \mathbb{R}^{n_x \times n_y} \) representing a two-dimensional visual stimulus of size \( n_x \times n_y \), the goal is to predict the corresponding outputs \( \{ \hat{\mathbf{I}}_1, \hat{\mathbf{I}}_2, \dots, \hat{\mathbf{I}}_{n_m} \} \), where each output \( \hat{\mathbf{I}}_i \) is a reconstruction of a past image from a specific time lag \( \tau \). To train the reservoir, we introduce a \emph{cognitive loss function}, denoted as \( \mathcal{L}_{\text{cog}} \), which encourages the network to preserve memory-related dynamics by minimizing the reconstruction error between the predicted image and its delayed ground truth counterpart. Specifically, the loss is defined as the average mean squared error (MSE) across all temporal lags:

\begin{equation}
\mathcal{L}_{\text{cog}} = \frac{1}{\tau_{\text{max}}} \sum_{\tau = 1}^{\tau_{\text{max}}} \frac{1}{T - \tau} \sum_{t = \tau + 1}^{T} \left\| \hat{\mathbf{I}}(t) - \mathbf{I}(t - \tau) \right\|^2
\end{equation}

Here, \( T \) is the total number of time steps, \( \tau_{\text{max}} \) is the maximum lag considered, and \( \| \cdot \| \) denotes the Frobenius norm. This formulation directly aligns the training objective with the preservation of delayed visual information, a key trait of memory.

To evaluate how effectively the reservoir retains visual information over time, we compute the \emph{Visual Memory Capacity (Vis-MC)}. This metric aggregates the squared Pearson correlation coefficient \( r_\tau^2 \) between the true input image at time \( t - \tau \) and the predicted image at time \( t \), over a range of delays:

\begin{equation}
\text{Vis-MC} = \sum_{\tau=1}^{\tau_{\text{max}}} r_\tau^2 ; r_\tau = \frac{\sum_{t=\tau+1}^{T} \left( \mathbf{I}(t-\tau) - \bar{\mathbf{I}}_\tau \right) \left( \hat{\mathbf{I}}(t) - \bar{\hat{\mathbf{I}}}_\tau \right)}{\sqrt{\sum_{t=\tau+1}^{T} \left\| \mathbf{I}(t-\tau) - \bar{\mathbf{I}}_\tau \right\|^2} \sqrt{\sum_{t=\tau+1}^{T} \left\| \hat{\mathbf{I}}(t) - \bar{\hat{\mathbf{I}}}_\tau \right\|^2}}
\end{equation}

Here, \( \bar{\mathbf{I}}_\tau \) and \( \bar{\hat{\mathbf{I}}}_\tau \) represent the mean input and predicted images, respectively, computed across time for each lag \( \tau \). A higher Vis-MC value indicates stronger visual memory retention and temporal alignment between past inputs and current predictions.

\subsubsection{Application to CBT Generation.} Our main application in this study is the generation of CBT, a population fingerprint at the network level. Specifically, we set out to learn a cognitively enhanced CBT from multi-view brain networks while adopting as base GNN architecture the deep graph normalizer (DGN) proposed in \cite{Gurbuz:2020}. 
DGN maps ROIs with identical attributes to high-dimensional distinctive representations by utilizing edge features \( e_{ij} \). This is achieved through \( l \) graph convolutional network layers with an edge-conditioned convolution operation \cite{simonovsky2017dynamic} separated by ReLU non-linearity. 
Each layer includes edge-conditioned filter learner network \( F^l \) that dynamically generates edge-specific weights for filtering message passing between ROIs \( p \) and \( q \) given the features of \( e_{pq} \). This operation is defined as:
\begin{equation} 
\mathbf{v}_p^l = \Theta^l  \mathbf{v}_p^{l-1} + \frac{1}{|\mathcal{N}(p)|} \left( \sum_{q \in \mathcal{N}(p)} F^l(e_{pq}; \mathbf{W}^l) \mathbf{v}_q^{l-1} + \mathbf{b}^l \right);  F^l(e_{pq}; \mathbf{W}^l) = \Theta_{pq}
\end{equation}
where \( \mathbf{v}_p^l \) is the embedding of ROI \( p \) at layer \( l \), \( \Theta^l \) is a learnable parameter, and \( \mathcal{N}(p) \) denotes the neighbors of ROI \( p \). \( \mathbf{b}^l \in \mathbb{R}^{d_l} \) denotes a network bias, and \( F^l \) is a neural network that maps \( \mathbb{R}^{n_v} \to \mathbb{R}^{d_l \times d_{l-1}} \) with weights \( \mathbf{W}^l \) where $d_{l-1}$ and $d_l$ represent input feature dimensions in layers $l-1$ and $l$, respectively. \( \Theta_{pq} \) represents the dynamically generated edge-specific weights by \( F^l \). Once the embeddings are learned, the CBT generation goes through two phases: the optimization phase and the refinement phase. 

\textsl{CBT optimization.}  
To generate the initial output CBT, we compute a set of tensor operations (Fig \ref{fig:main}.B1) on the learned embeddings $V$ of the final layer $L$ for easy and efficient backpropagation. First, \(V\) is replicated horizontally \( n_r \) times to obtain \( \mathcal{R} \in \mathbb{R}^{n_r \times n_r \times d_L} \). Next, \( \mathcal{R} \) is transposed to get \( \mathcal{R}^\top \). Lastly, we compute the element-wise absolute difference of \( \mathcal{R} \) and \( \mathcal{R}^\top \). The resulting tensor is summed along the \( z \)-axis to estimate the CBT \( \mathbf{C} \in \mathbb{R}^{n_r \times n_r} \) \cite{Gurbuz:2020}. To enhance the centeredness of $\mathbf{C}$, we use a randomly sampled subset of the training subjects (Fig \ref{fig:main}.B1). This approach offers two key advantages: (1) it regularizes the model by reducing overfitting, and (2) allows a fixed sample size, ensuring the loss magnitude and computation time remain independent of the training set size \cite{Gurbuz:2020}. Given the generated CBT \( \mathbf{C}_s \) for subject \( s \) and a random subset \( R \) of training subject indices, we define the centeredness loss function as:

\begin{equation}
    \mathcal{L}_{gnn} = \mathcal{L}_{C_s} = \sum_{v=1}^{n_v} \sum_{i \in R} \| \mathbf{C}_s - \mathcal{T}^v_i \lambda_v \|_F,
    \quad \min_{C_s} \frac{1}{n_s} \sum_{s=1}^{n_s} \mathcal{L}_{C_s}
\end{equation}

\noindent where \( n_s \) is the number of subjects, \( \| \cdot \|_F \) is the Frobenius norm, \(\mathcal{T}^v_i \in \mathbb{R}^{n_r \times n_r}\) is the \(v\)-th view of subject \(i\) and \( \lambda_v \) is a view-specific normalizer defined as
$ \lambda_v = {\frac{1}{\mu_v}} /{\max \bigg\{ \frac{1}{\mu_j} \bigg\}_{j=1}^{n_v}} $
where $ \mu_v$ denotes the average weights of view $v$. This view-specific normalization helps avoid view-biased CBT estimation since brain network connectivity distribution and value range might largely vary across views \cite{Gurbuz:2020}.

To enhance the cognitive awareness of the generated templates, we propose a \emph{co-optimization process} that integrates both cognitive loss and centeredness loss. This ensures the reservoir learns to recall spatial and intensity patterns from input image sequences, while enforcing that the CBT remains representative of the population. The framework alternates between two objectives: 1) \emph{CBT Optimization:} Update the CBT \( \mathbf{C}_s \) using the centeredness loss \( \mathcal{L}_{C_s} \) to maintain population representativeness. 2) \emph{Reservoir Optimization:} Use the updated \( \mathbf{C}_s \) as the reservoir input to optimize the readout weights \( \mathbf{W}_{\text{out}} \) via cognitive loss \( \mathcal{L}_{\text{cog}} \), enhancing visual memory recall. The co-optimization objective is:

\begin{equation}
\min_{\mathbf{W}_{\text{out}}, \mathbf{C}_s} 
\underbrace{\frac{1}{n_s}\sum_{s=1}^{n_s}\mathcal{L}_{C_s}}_{\text{centeredness loss}} +
\underbrace{\frac{1}{\tau_{\text{max}}(T - \tau)} \sum_{\tau=1}^{\tau_{\text{max}}} \sum_{t = \tau + 1}^{T} \|\mathbf{I}(t - \tau) - \hat{\mathbf{I}}(t)\|_2^2}_{\text{cognitive loss } (\mathcal{L}_{\text{cog}})}
\end{equation}

where the predicted image \(\hat{\mathbf{I}} = \mathbf{W}_{\text{out}} \mathbf{x}_i\), and \(\mathbf{x}_i\) are the reservoir states \(\hat{\mathbf{x}}_i\) (Eq.~\eqref{eq:state_update}) driven by the cognitive reservoir. This iterative process enhances both the CBT's structural centeredness and the reservoir’s cognitive prediction ability, creating a dynamic feedback loop between CBT generation and reservoir training.

\textsl{CBT refinement.} While each subject's multi-view brain network is mapped to a CBT \( \mathbf{C}_s \), these may reflect subject-specific biases. To correct this, we adopt a refinement step (Fig. \ref{fig:main}.B2) \cite{Gurbuz:2020}, aggregating all individual CBTs in the training set using an element-wise median. This yields a final population-level CBT that is symmetric, non-negative, and more representative of the group.

\section{Experiments and Results}
\textbf{Evaluation datasets.} We evaluate our model on two datasets. The first (AD/LMCI) is from the Alzheimer’s Disease Neuroimaging Initiative GO dataset \cite{Mueller2005}, comprising 77 individuals (41 AD, 36 LMCI) from both hemispheres. Each subject is described by four cortical morphological attributes: sulcal depth, mean cortical thickness, maximum principal curvature, and average curvature. These are extracted from T1-weighted MRI scans using the FreeSurfer pipeline \cite{fischl2012freesurfer} and segmented into 35 ROIs via the Desikan-Killiany atlas \cite{desikan2006automated}, with brain networks built from pairwise absolute differences. The second dataset (ASD/NC) is processed similarly, using data from the ABIDE I dataset \cite{di2014autism}, with 310 individuals (155 NC, 155 ASD). It includes additional features such as cortical surface area and minimum principal curvature, yielding six connectional features per subject. Left and right hemispheres are analyzed separately in both datasets.

\noindent\textbf{Hyperparameter tuning and training.} We train separate models for left (LH) and right hemispheres (RH) across four groups, each using three edge-conditioned convolution layers with ReLU-activated neural networks. These layers learn dynamic message-passing filters \( \Theta_{ij}^l \) to map features from heterogeneous views. The resulting node embeddings have dimensions 36--5 (AD/LMCI) and 8 (NC/ASD). Models are trained using 5-fold cross-validation (80/20 split), Adam optimizer (500 epochs, \( lr = 0.005 \)), and early stopping to prevent overfitting.

\noindent\textbf{Visual memory recall task.} For the visual memory recall task, we use the MNIST dataset \cite{deng2012mnist} as input to the ESN. MNIST’s simplicity and standardized format enable consistent evaluation, while its rich spatial features, such as varying digit shapes, make it suitable for testing visual memory recall. To reduce computational complexity, the images are resized to \( 10 \times 10 \) pixels using mean pooling. At each time step, 15 images are used for training and 5 for testing, with time lags ranging from 5 to 40. The ESN is configured with the following parameters: spectral radius \( \rho = 0.98 \), input scaling \( \epsilon = 1 \times 10^{-6} \), leakage rate \( \alpha = 1 \), bias \( b = 0 \), and transient steps \( n_{\text{transient}} = 100 \). The input weight matrix \( \mathbf{W}_{\text{in}} \) is derived from a uniform distribution with values in the interval \( ]-1, 1[ \). The ESNs are implemented using the \texttt{echoes} library \cite{damicelli2019echoes}.

\noindent\textbf{Benchmark method.} We selected DGN \cite{Gurbuz:2020} as our benchmark due to its strong performance in CBT generation. DGN outperformed seven leading methods as reported in the comparative survey \cite{ChaariAkdagRekik2022MultigraphIntegration}, making it a relevant and rigorous baseline. However, DGN lacks cognitive components, which are central to our approach.


\noindent\textbf{CBT evaluation.} We evaluate the generated CBTs based on centeredness and visual capacity, and report additional analyses—topology, discriminative power, key biomarkers, and CBT visualizations—in the \href{supplementary.pdf}{supplementary material}.

\begin{figure}[ht!]
    \centering
    \includegraphics[width=0.87\textwidth]{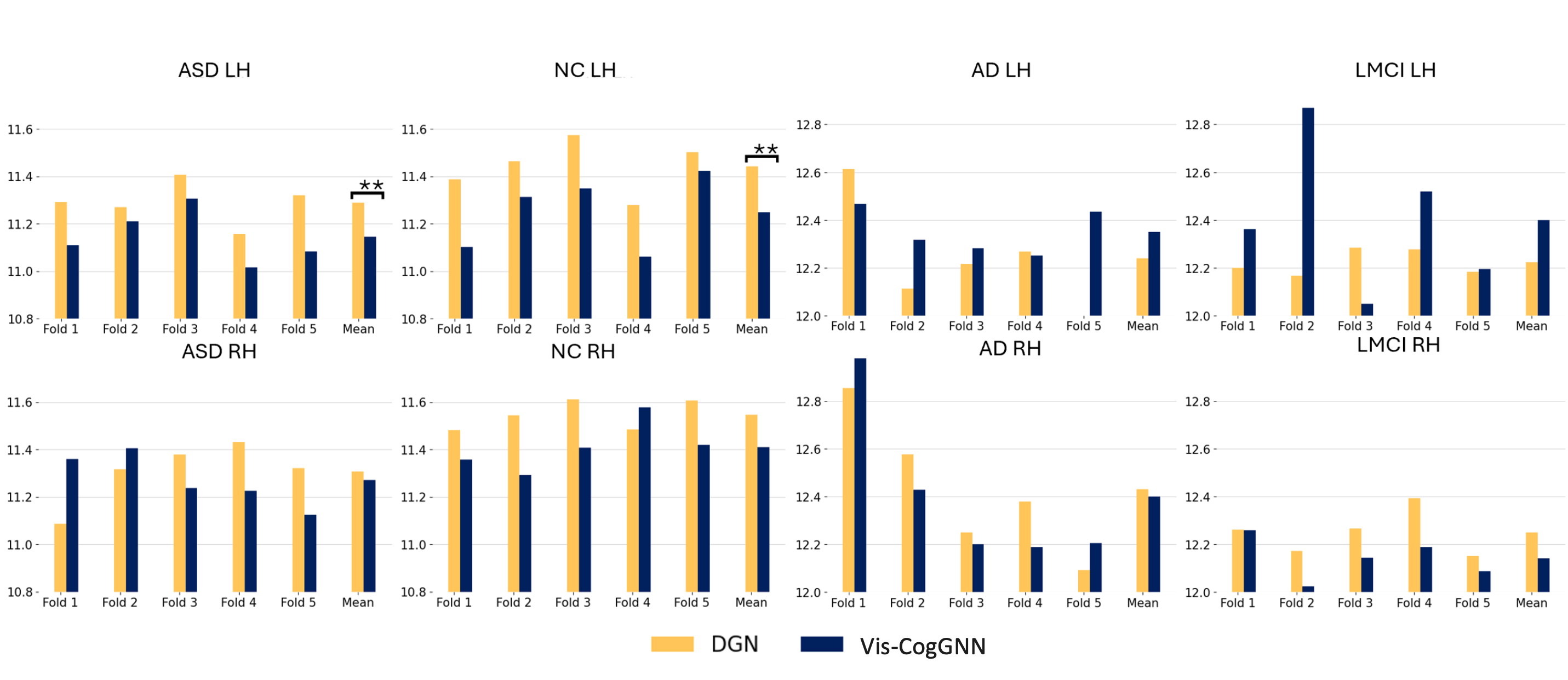}
\caption{\textsl{Centeredness comparison.} (**) indicates \( p < 0.005 \), two-tailed paired \( t \)-test.}

    \label{fig:centeredness}
\end{figure}

\noindent{\emph{Centeredness.}} To assess the centeredness of the generated CBTs, we compute the mean Frobenius distance $d_F(\mathbf{C}, \mathcal{A}) = \frac{1}{n_v} \sum_{k=1}^{n_v} \| \mathbf{C} - \mathcal{A}_k \|_F $
which measures the average deviation between the CBT \( \mathbf{C} \) and the individual network views \( \mathcal{A}_k \) in the testing set. A lower Frobenius distance indicates better centeredness, meaning the CBT is more representative and closely aligned with individual subject networks. We split both datasets using 5-fold cross-validation, generating CBTs from the training folds and evaluating centeredness on the held-out test folds. Our results show that \submodelname\ achieves lower Frobenius distances than DGN in 75\% of cases (Fig. \ref{fig:centeredness}), reflecting the benefit of incorporating cognitive capacities into the template generation. The reduced centeredness in LH for AD and LMCI groups likely reflects greater variability in connectivity due to neurodegeneration. While RH shows pronounced asymmetries linked to visuospatial and executive deficits \cite{kannappan2018asymmetrical}, the LH exhibits subtler but widespread disruptions, making it harder to generate a consistent, well-aligned CBT. 

\begin{figure}[ht!]
    \centering
    \includegraphics[width=0.88\textwidth]{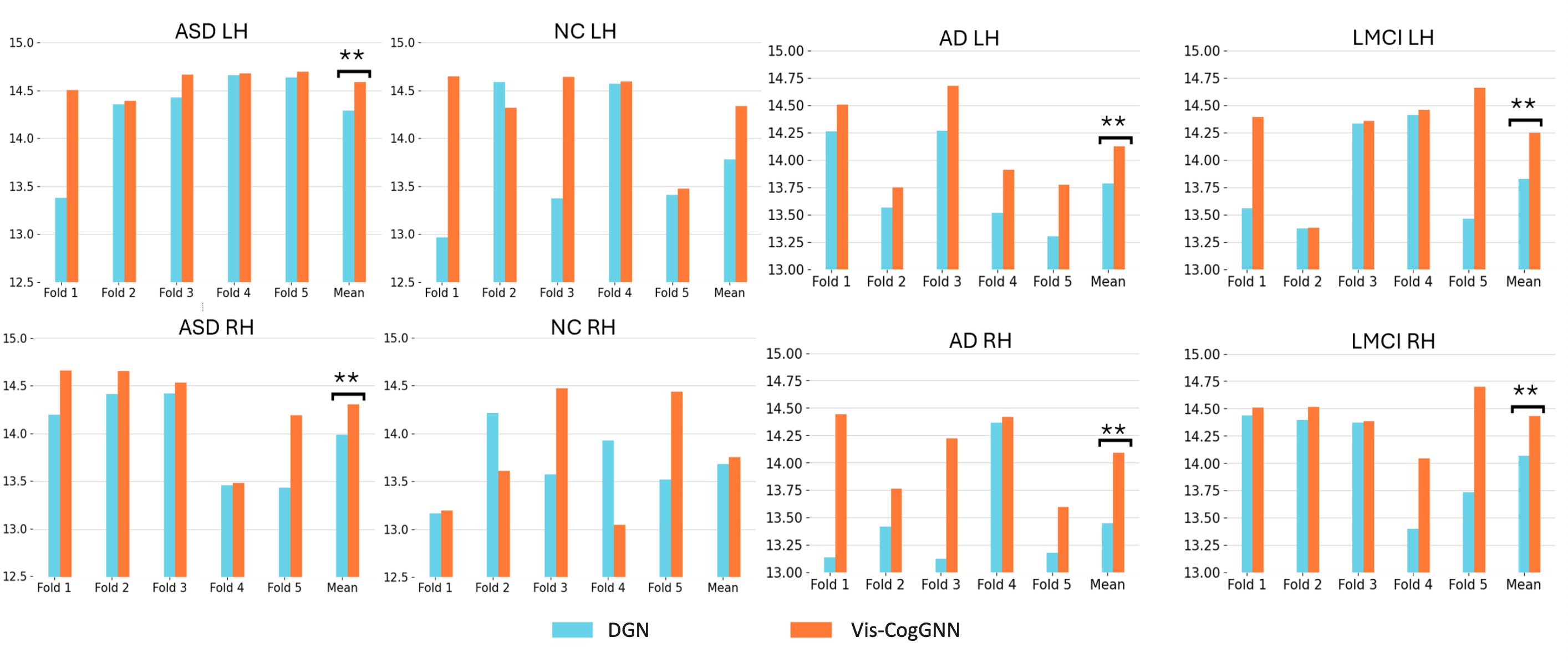}
    \caption{\textsl{Vis-MC comparison between} \textsc{\submodelname} \textsl{and DGN \cite{Gurbuz:2020}. (\(**\) for \(p < .005)\)}
}
    \label{fig:spatial_memory}
\end{figure}

\noindent{\emph{Visual memory capacity.}} The visual memory evaluation measures how well the generated templates retain cognitive features by reconstructing the visual input (MNIST images). We compare CBTs generated by DGN and \textsc{\submodelname} using the Vis-MC score (Section~2.2), where higher values indicate stronger cognitive capacity. As shown in Fig. \ref{fig:spatial_memory}, \textsc{\submodelname} significantly outperforms DGN, achieving statistical significance (\(p < 0.005\)) in six out of eight datasets. This demonstrates its superior ability to preserve cognitive traits, especially in complex cases like AD-LH and LMCI-LH. \textsc{\submodelname} also shows consistent performance across folds, unlike DGN, which is more sensitive to sample variation. These results underscore the benefit of integrating cognitive mechanisms into CBT generation, a key feature of our approach.

\section{Conclusion}
We introduced \textsc{\modelname}, the first \emph{cognitive generative} GNN model, designed for broad applicability across learning tasks. As a case study, we proposed \textsc{\submodelname}, a specialized variant that integrates visual input and uses brain connectomes as a reservoir to enhance visual memory recall. Applied to CBT generation, our model produces templates that preserve cognitive features. Future work will explore broader applications and richer cognitive inputs to expand its utility in network neuroscience and beyond.

\clearpage
\appendix
\begin{center}
  \LARGE \textcolor{blue}{Supplementary Material}: \modelname: Cognitive Graph Neural Networks in Generative Connectomics
\end{center}
\vspace{1em}

\renewcommand{\thefigure}{S\arabic{figure}}
\renewcommand{\thetable}{S\arabic{table}}
\setcounter{figure}{0}
\setcounter{table}{0}

\begin{figure}[ht!]
    \centering
    \includegraphics[width=1\linewidth]{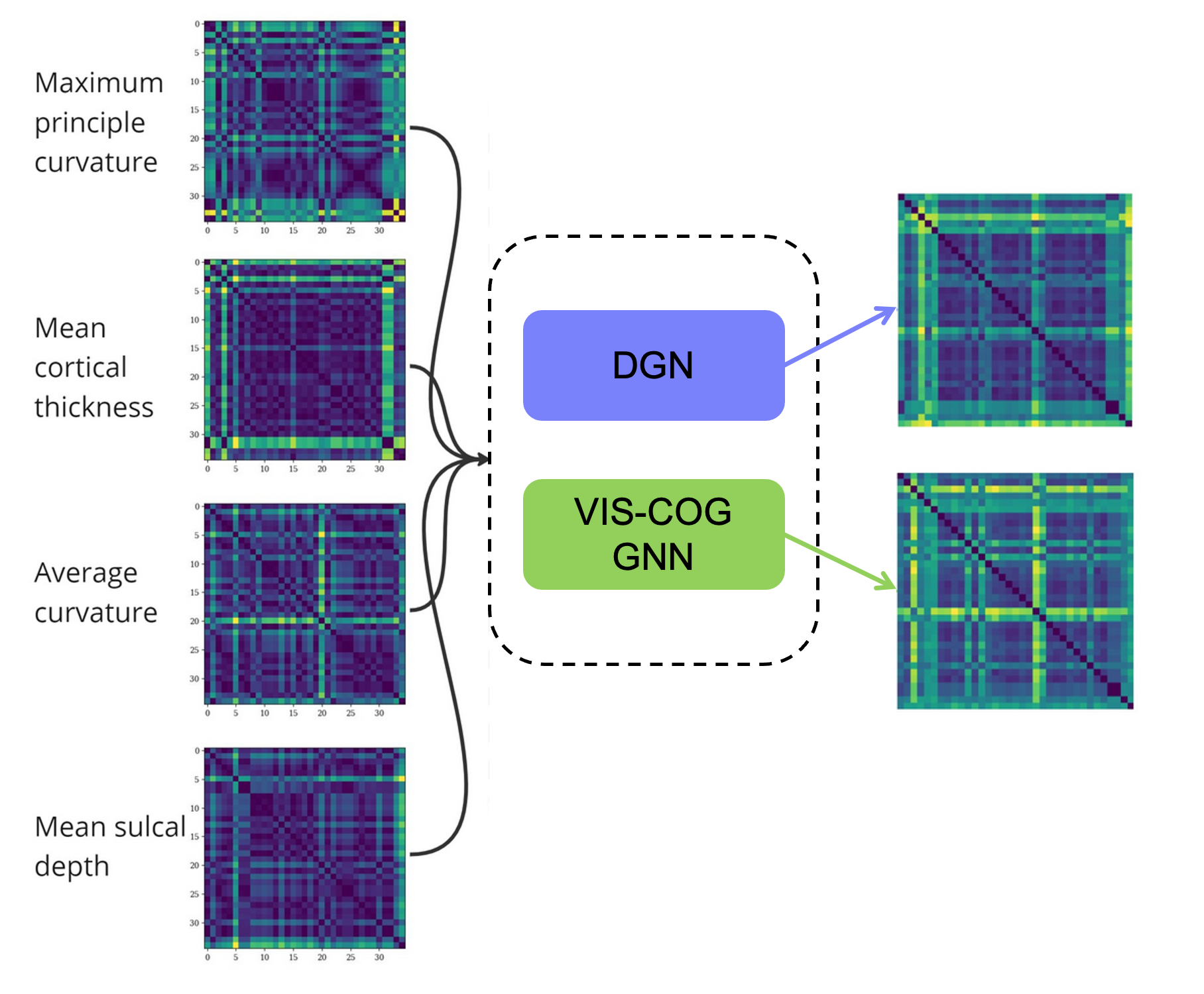}
    \caption{Example CBTs generated by DGN \cite{Gurbuz:2020} and \submodelname for the AD group (left hemisphere), derived from four brain structural attributes: maximum principal curvature, mean cortical thickness, average curvature, and mean sulcal depth. The left panel shows individual view-specific brain networks, while the right panel presents the resulting population-level templates. Both methods produce symmetric CBTs, but \submodelname generates richer and more distinct patterns by integrating cognitive information during the generative process.}

    \label{fig:cbts}
\end{figure}

Here we visualize the generated CBTs in Fig.~\ref{fig:cbts} to illustrate the patterns captured by each method. We then report the discriminative power, topological soundness, and highlight the most discriminative features identified.\\

\noindent\textbf{\emph{CBT Visualization.}} To qualitatively compare the patterns captured by different models, we visualize example CBTs in Fig. \ref{fig:cbts}. The left panel shows individual subject-level brain networks derived from four structural attributes: \emph{maximum principal curvature}, \emph{mean cortical thickness}, \emph{average curvature}, and \emph{mean sulcal depth}, from the left hemisphere of the AD group. These view-specific networks are then integrated by the baseline DGN \cite{Gurbuz:2020} and the proposed \submodelname. The right panel displays the resulting population-level templates produced by each method. While both approaches yield symmetric templates, \submodelname generates \emph{richer and more differentiated connectivity patterns}, reflecting its ability to incorporate cognitive information into the template estimation process.

\begin{table}[ht!]
\centering
\caption{Discriminative performance of SVM-DGN and SVM-\submodelname across multiple datasets. 
SVM-DGN refers to an SVM classifier trained on CBTs generated by DGN \cite{Gurbuz:2020}; 
SVM-\submodelname is trained on CBTs from our method.}
\label{tab:discriminative_performance}
\setlength{\tabcolsep}{6pt}
\begin{tabular}{lcccccc}
\hline
\textbf{Dataset} & \textbf{Method} & \textbf{ACC} & \textbf{SS} & \textbf{SP} & \textbf{AUC} & \textbf{F1} \\ \hline

\multirow{2}{*}{\textbf{AD/LMCI LH}} 
    & SVM-DGN         & 55.32 & 80.20 & 30.44 & 55.87 & 64.17 \\ 
    & SVM-\submodelname & \textbf{66.00} & \textbf{100.00} & \textbf{32.00} & \textbf{69.76} & \textbf{74.63} \\ \hline

\multirow{2}{*}{\textbf{AD/LMCI RH}} 
    & SVM-DGN         & 51.67 & 39.26 & 64.07 & 52.47 & 35.33 \\ 
    & SVM-\submodelname & \textbf{74.00} & \textbf{48.00} & \textbf{100.00} & \textbf{87.68} & \textbf{64.86} \\ \hline

\multirow{2}{*}{\textbf{ASD/NC LH}} 
    & SVM-DGN         & 52.49 & 34.58 & \textbf{70.40} & 54.44 & 37.66 \\ 
    & SVM-\submodelname & \textbf{68.00} & \textbf{100.00} & 36.00 & \textbf{84.32} & \textbf{71.43} \\ \hline

\multirow{2}{*}{\textbf{ASD/NC RH}} 
    & SVM-DGN         & 51.39 & 22.33 & \textbf{80.47} & 55.93 & 24.10 \\ 
    & SVM-\submodelname & \textbf{68.00} & \textbf{100.00} & 36.00 & \textbf{84.32} & \textbf{75.76} \\ \hline

\end{tabular}
\end{table}

\noindent\textbf{\emph{CBT discriminativeness.}} To evaluate the discriminative power of the generated CBTs, we use a CBT-shot learning approach with a support vector machine (SVM) classifier. The SVM is trained on CBTs from both classes (e.g., AD and LMCI) and assessed using 5-fold cross-validation. Key metrics include accuracy (ACC), sensitivity (SS), specificity (SP), area under the curve (AUC), and F1 score, as reported in Table~\ref{tab:discriminative_performance}. \textsc{\submodelname} consistently outperforms DGN \cite{Gurbuz:2020}, particularly in accuracy, sensitivity, and AUC, demonstrating stronger class-separability. These gains are especially pronounced in neurodegenerative datasets (AD/LMCI), where \textsc{\submodelname} achieves up to 22\% higher accuracy and nearly perfect sensitivity in the left hemisphere. Notably, it also maintains strong performance on ASD/NC classification, with improvements in both F1 and AUC. This suggests that integrating cognitive signals into the generative process leads to more informative and discriminative templates, enabling more robust subject-level classification even with limited supervision.\\

\begin{figure}[ht!]
    \centering
    \includegraphics[width=1\linewidth]{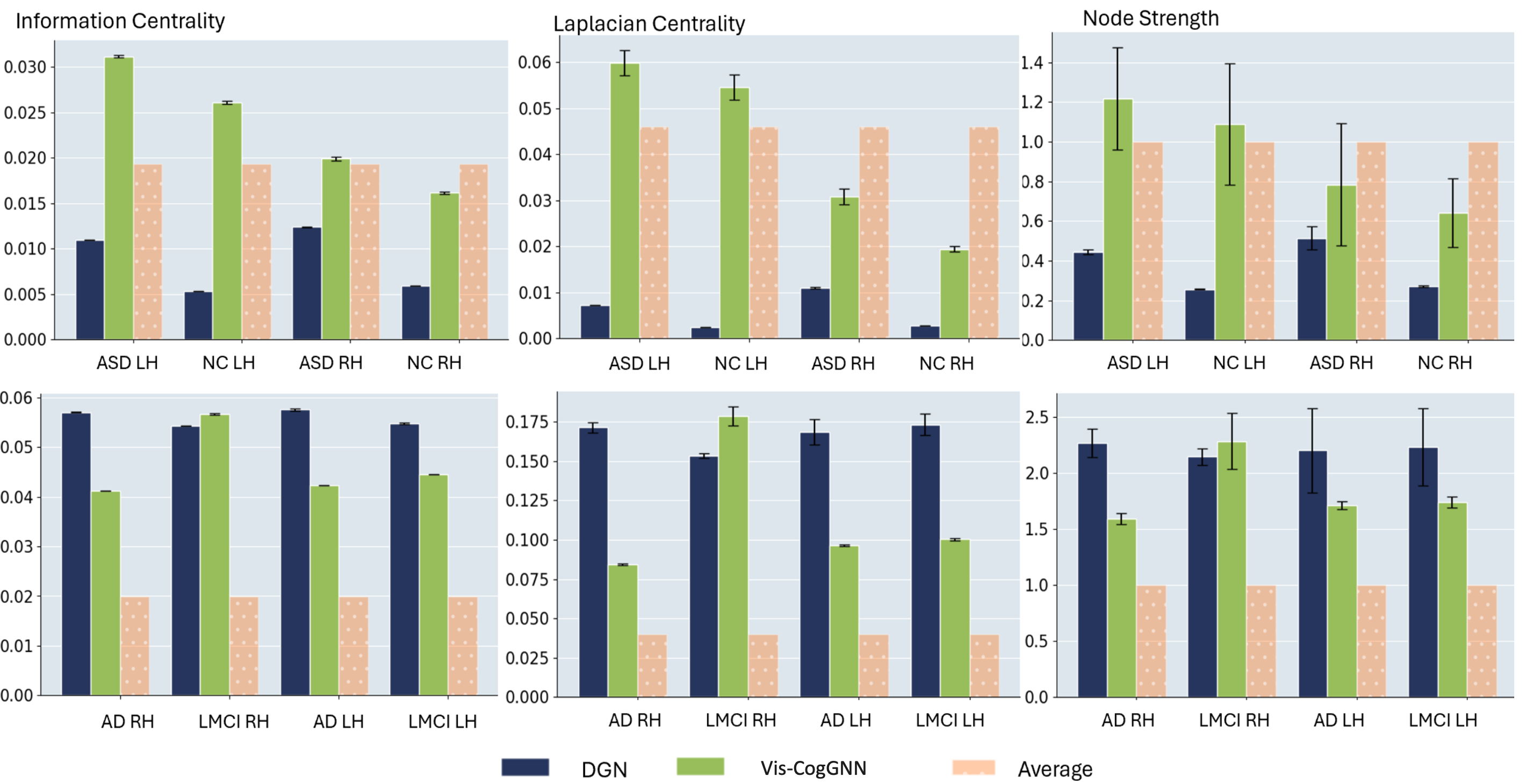}
    \caption{\textsl{Topological comparison}. The charts display the average distributions of information centrality, Laplacian centrality, and node strength for CBTs estimated using DGN \cite{Gurbuz:2020} and \submodelname, alongside reference values from dataset averages. Mean and variance are shown for left and right hemispheres across NC, ASD, AD, and LMCI datasets.
}
    \label{fig:topology}
\end{figure}

\noindent\textbf{\emph{Topological Soundness.}} We assessed the topological soundness of the generated CBTs by comparing their structural properties---\textsl{information centrality}, \textsl{Laplacian centrality}, and \textsl{node strength}---against the average connectivity profile of the population (used here as a reference baseline rather than a ground truth). As shown in Fig.~\ref{fig:topology}, \submodelname consistently produces higher values than DGN \cite{Gurbuz:2020} across all three metrics in the ASD and NC datasets, reflecting a richer and more well-structured topology. These differences are statistically significant (two-tailed paired \( t \)-test), indicating that \submodelname better preserves meaningful topological traits in healthy and neurodiverse brain networks. For the AD and LMCI datasets, both methods exhibit lower metric values overall, with Vis-CogGNN closely matching the population average. This suggests that in the presence of neurodegeneration, characterized by widespread disruptions in connectivity, \submodelname generates simpler, more representative templates, aligned with the reduced topological complexity typical of these conditions. Together, these results highlight \submodelname’s ability to adapt its generative modeling to the underlying population structure, preserving topological richness when present and simplifying appropriately in disease settings. This invites further exploration of how cognitive-informed modeling interacts with brain network topology across health and pathology. \\

\noindent\textbf{\emph{Biomarker discovery}.} To further evaluate the discriminative power of the generated CBTs using \submodelname, we identified the top discriminative connections from the learned templates (Table \ref{tab:discriminative_connections}). For ASD, the key discriminative connections involve the \textsl{caudal middle frontal gyrus} (CMFG) and \textsl{cuneus cortex} (CC). The CMFG, known for its role in executive functions such as planning and flexibility, shows consistent disruptions, reflecting its involvement in the cognitive deficits observed in ASD ~\cite{philip2012systematic}. Similarly, the CC, associated with visual processing, demonstrates alterations that align with the atypical visual and spatial processing frequently reported in ASD~\cite{zhou2023children}. For AD,  significant connections involve regions such as \textsl{the bank of the superior temporal sulcus (BSTS)}, \textsl{caudal middle frontal gyrus (CMFG)}, and \textsl{precuneus cortex} (PC). Degeneration in the BSTS affects social cognition and sensory integration, while atrophy in the CMFG impairs executive functions like decision-making and working memory~\cite{bertoux2016social}. Finally, The PC is associated with episodic memory and spatial orientation, with alterations contributing to memory loss and disorientation in AD ~\cite{bailly2015precuneus}.

\vspace{-1em} 
\begin{table}[!ht]
\caption{Top five discriminative connections identified for each population pair}
\label{tab:discriminative_connections}
\centering
\resizebox{1\textwidth}{!}{ 
\begin{tabular}{|c|c|c|}
\hline
\textbf{Rank} & \textbf{NC-ASD RH} & \textbf{NC-ASD LH} \\ \hline
1 & Medial orbital frontal cortex $\leftrightarrow$ Isthmus-cingulate cortex & Caudal middle frontal gyrus $\leftrightarrow$ Entorhinal cortex \\ \hline
2 & Pars orbitalis $\leftrightarrow$ Pericalcarine cortex & Caudal middle frontal gyrus $\leftrightarrow$ Pericalcarine cortex \\ \hline
3 & Pars triangularis $\leftrightarrow$ Precuneus cortex & Caudal middle frontal gyrus $\leftrightarrow$ Cuneus cortex \\ \hline
4 & Bank of the superior temporal sulcus $\leftrightarrow$ Middle temporal gyrus & Cuneus cortex $\leftrightarrow$ Bank of the superior temporal sulcus \\ \hline
5 & Caudal middle frontal gyrus $\leftrightarrow$ Paracentral lobule & Entorhinal cortex $\leftrightarrow$ Paracentral lobule \\ \hline
\end{tabular}
}

\resizebox{1\textwidth}{!}{
\begin{tabular}{|c|c|c|}
\hline
\textbf{Rank} & \textbf{AD-LMCI RH} & \textbf{AD-LMCI LH} \\ \hline
1 & \textbf{Entorhinal cortex $\leftrightarrow$ Caudal middle frontal gyrus} & Inferior parietal cortex $\leftrightarrow$ Precuneus cortex \\ \hline
2 & Caudal middle frontal gyrus $\leftrightarrow$ Lingual gyrus & Cuneus cortex $\leftrightarrow$ Supramarginal gyrus \\ \hline
3 & Transverse temporal cortex $\leftrightarrow$ Superior parietal cortex & Bank of the superior temporal sulcus $\leftrightarrow$ Temporal pole \\ \hline
4 & Unmeasured Corpus Callosum $\leftrightarrow$ Transverse temporal cortex & Bank of the superior temporal sulcus $\leftrightarrow$ Insula cortex \\ \hline
5 & Medial orbital frontal cortex $\leftrightarrow$ Bank of the superior temporal sulcus & Rostral middle frontal gyrus $\leftrightarrow$ Precuneus cortex \\ \hline
\end{tabular}
}

\end{table}

\bibliographystyle{splncs}
\bibliography{Biblio}

\end{document}